\documentclass[11pt,a4paper]{article}
\usepackage[hyperref]{acl2020}
\usepackage{times}
\usepackage{latexsym}

\usepackage{microtype}

\usepackage{todonotes}
\usepackage{ulem}
\usepackage{multirow}
\usepackage{amssymb}
\usepackage{paralist}
\usepackage{amsmath}
\usepackage{soul}
\usepackage{booktabs}
\usepackage{subfigure}
\usepackage{color}
\usepackage{colortbl}
\usepackage{nicefrac}

\DeclareMathOperator*{\argmax}{arg\,max}
\newcommand{\dataset}{MLQA}
\newcommand{\papername}{\dataset{}: Evaluating Cross-lingual Extractive Question Answering}

\usepackage{url}

\aclfinalcopy % Uncomment this line for the final submission
\def\aclpaperid{203} %  Enter the acl Paper ID here

\title{\papername{}}

\author{
  Patrick Lewis$^{\text{*}\dagger{}}$ \hspace{0.15cm} 
  Barlas O\u{g}uz$^{\text{*}}$ \hspace{0.15cm} 
  Ruty Rinott$^{\text{*}}$ \hspace{0.15cm} 
  Sebastian Riedel$^{\text{*}\dagger{}}$ \hspace{0.15cm} 
  Holger Schwenk$^{\text{*}}$ \\
  $^{\text{*}}$Facebook AI Research \hspace{0.3cm} $^{\dagger{}}$University College London \\
  \texttt{\{plewis,barlaso,ruty,sriedel,schwenk\}@fb.com}
 }
\date{}
\begin{document}
\normalem

\maketitle
\begin{abstract}

Question answering (QA) models have shown rapid progress enabled by the availability of large, high-quality benchmark datasets. 
Such annotated datasets are difficult and costly to collect, and rarely exist in languages other than English, making building QA systems that work well in other languages challenging. 
In order to develop such systems, it is crucial to invest in high quality multilingual \emph{evaluation} benchmarks to measure progress. 
We present \dataset{}, a multi-way aligned extractive QA evaluation benchmark intended to spur research in this area.\footnote{\dataset{} is publicly available at \url{https://github.com/facebookresearch/mlqa}} \dataset{} contains QA
instances in 7 languages, \emph{English, Arabic, German, Spanish, Hindi, Vietnamese} and \emph{Simplified Chinese}.
\dataset{} has over 12K instances in English and 5K in each other language, with each instance parallel between 4 languages on average.
We evaluate state-of-the-art cross-lingual models and machine-translation-based baselines
on \dataset{}.  In all cases, transfer results are significantly behind training-language performance.

\end{abstract}

\section{Introduction}

Question answering (QA) is a central and highly popular area in NLP, with an abundance of datasets available to tackle the problem from various angles, including extractive QA, cloze-completion, and open-domain QA~\cite{richardson_mctest:_2013,rajpurkar_squad:_2016,chen_reading_2017,kwiatkowski_natural_2019}. The field has made rapid advances in recent years, even exceeding human performance in some settings ~\cite{devlin_bert:_2019,alberti_synthetic_2019}.

Despite such popularity, QA datasets in languages other than English remain scarce, even for relatively high-resource languages \cite{asai_multilingual_2018}, as collecting such datasets at sufficient scale and quality is difficult and costly.  There are two reasons why this lack of data prevents internationalization of QA systems. First, we cannot \emph{measure} progress on multilingual QA without relevant benchmark data. Second, we cannot easily \emph{train} end-to-end QA models on the task, and arguably most recent successes in QA have been in fully supervised settings. Given recent progress in cross-lingual tasks such as  document classification \cite{Lewis:Reuters:2004,klementiev_inducing_2012,Schwenk:2018:lrec_mldoc}, semantic role labelling~\cite{akbik_generating_2015} and NLI~\cite{conneau_xnli:_2018}, we argue that while multilingual QA training data might be useful but not strictly necessary, multilingual evaluation data is a must-have.

Recognising this need, several cross-lingual datasets have recently been assembled~\cite{asai_multilingual_2018,liu_xqa:_2019}. 
However, these generally cover only a small number of languages, combine data from different authors and annotation protocols, lack parallel instances, or explore less practically-useful QA domains or tasks (see Section \ref{related}).
Highly parallel data is particularly attractive, as it enables fairer comparison across languages, requires fewer source language annotations, and allows for additional evaluation setups at no extra annotation cost. A purpose-built evaluation benchmark dataset covering a range of diverse languages, and following the popular extractive QA paradigm on a practically-useful domain would be a powerful testbed for cross-lingual QA models.

With this work, we present such a benchmark, \dataset{}, and hope that it serves as an accelerator for multilingual QA in the way datasets such as SQuAD \cite{rajpurkar_squad:_2016} have done for its monolingual counterpart.
\dataset{} is a multi-way parallel extractive QA evaluation benchmark in seven languages:
\textit{English, Arabic, German, Vietnamese, Spanish, Simplified Chinese} and \textit{Hindi}.
To construct \dataset{}, we first automatically identify sentences from Wikipedia articles which have the same or similar meaning
in multiple languages.  
We extract the paragraphs that contain such sentences, then crowd-source questions on the English paragraphs, making sure the answer is in the aligned sentence.
This makes it possible to answer the question in all languages in the vast majority of cases.\footnote{The automatically aligned sentences occasionally differ in a named entity or information content, or some questions may not make sense without the surrounding context.  In these rare cases, there may be no answer for some languages.} 
The generated questions are then translated to all target languages by professional translators, and answer spans are annotated in the aligned contexts for the target languages. 

The resulting corpus has between 5,000 and 6,000 instances in each language, and more than 12,000 in English. Each instance has an aligned equivalent in multiple other languages (always including English), the majority being 4-way aligned. Combined, there are over 46,000 QA annotations. 

We define two tasks to assess performance on \dataset{}. The first, cross-lingual transfer~(XLT), requires models trained in one language (in our case English) to transfer to test data in a different language. The second, generalised cross-lingual transfer~(G-XLT) requires models to answer questions where the question and context language is \emph{different},
e.g. questions in Hindi and contexts in Arabic, a setting possible because \dataset{} is highly parallel.

We provide baselines using state-of-the-art cross-lingual techniques. We develop machine translation baselines which map answer spans based on the attention matrices from a translation model, and use multilingual BERT \cite{devlin_bert:_2019} and XLM \cite{lample_cross-lingual_2019} as zero-shot approaches.
We use English for our training language and adopt SQuAD as a training dataset. We find that zero-shot XLM transfers best, but all models lag well behind training-language performance.

In summary, we make the following contributions: \begin{inparaenum}[i)]
    \item We develop a novel annotation pipeline to construct large multilingual, highly-parallel extractive QA datasets
    \item We release \dataset{}, a 7-language evaluation dataset for cross-lingual QA
    \item We define two cross-lingual QA tasks, including a novel generalised cross-lingual QA task
    \item We provide baselines using state-of-the-art techniques, and demonstrate significant room for improvement.
\end{inparaenum}

%----------------------------------------------------
\section{The \dataset{} corpus}\label{dataset}

First, we state our desired properties for a cross-lingual QA evaluation dataset. We note that whilst some existing datasets exhibit these properties, none exhibit them all in combination (see section \ref{related}). We then describe our annotation protocol, which seeks to fulfil these desiderata. 

\paragraph{Parallel}
The dataset should consist of instances that are parallel across many languages. First, this makes comparison of QA performance as a function of transfer language fairer. Second, additional evaluation setups become possible, as questions in one language can be applied to documents in another.
Finally, annotation cost is also reduced as more instances can be shared between languages.

\paragraph{Natural Documents}
Building a parallel QA dataset in many languages requires access to parallel documents in those languages. Manually translating documents at sufficient scale entails huge translator workloads, and could result in unnatural documents. 
Exploiting existing naturally-parallel documents is advantageous, providing high-quality documents without requiring manual translation.

\paragraph{Diverse Languages}
A primary goal of cross-lingual research is to develop systems that work well in many languages. The dataset should enable quantitative performance comparison across languages with different linguistic resources, language families and scripts.

\paragraph{Extractive QA} Cross-lingual understanding benchmarks are typically based on classification~\cite{conneau_xnli:_2018}.
Extracting spans in different languages represents a different language understanding challenge.
Whilst there are extractive QA datasets in a number of languages (see Section~\ref{related}), most were created at different times by different authors with different annotation setups, making cross-language analysis challenging.

\paragraph{Textual Domain}
We require a naturally highly language-parallel textual domain. Also, it is desirable to select a textual domain that matches existing extractive QA training resources, in order to isolate the change in performance due to language transfer.

\begin{figure*}
\centering

  \includegraphics[width=\textwidth]{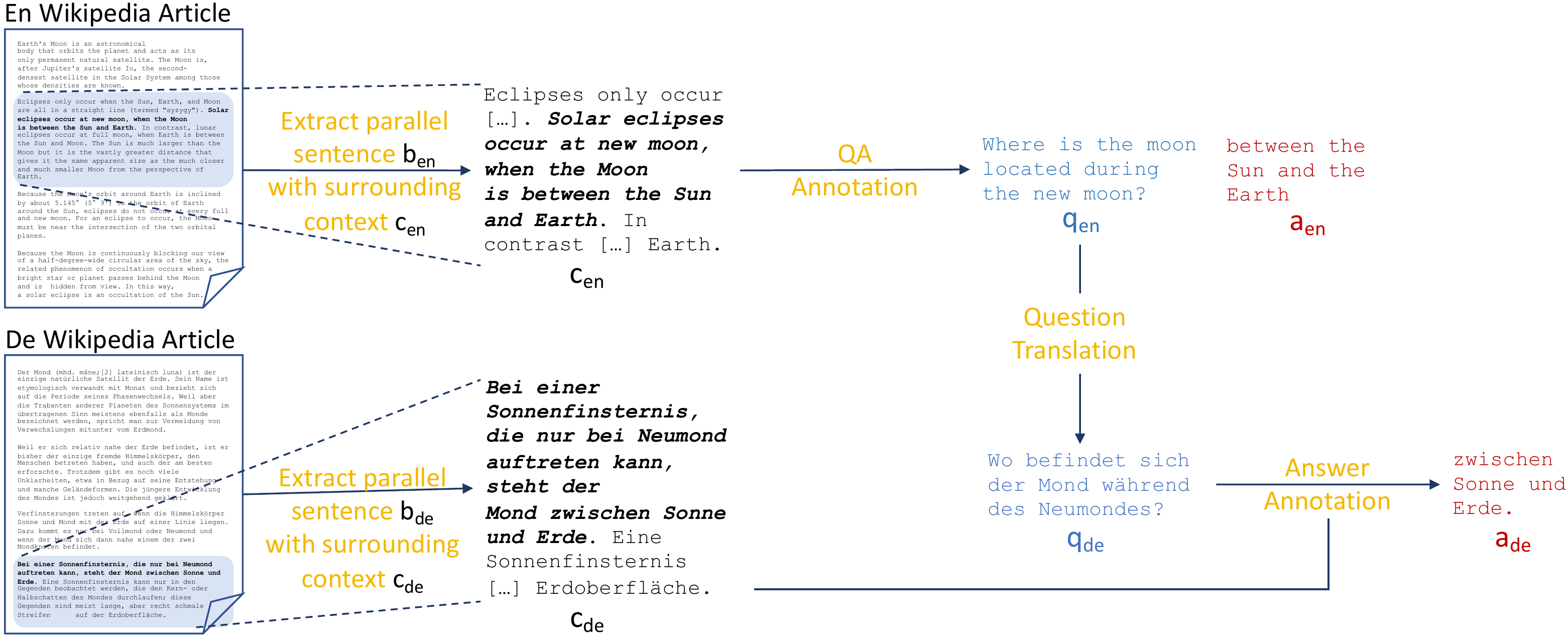}
\caption{
    \dataset{} annotation pipeline. Only one target language is shown for clarity.
    Left: We first identify $N$-way parallel sentences $b_{en}$, $b_1$ \ldots $b_{N-1}$ in Wikipedia articles on the same topic, and extract the paragraphs that contain them, $c_{en}$, $c_1$ \ldots $c_{N-1}$.
    Middle: Workers formulate questions $q_{en}$ from $c_{en}$ for which answer $a_{en}$ is a span within $b_{en}$.
    Right: English questions $q_{en}$ are then translated by professional translators into all languages $q_i$ and the answer $a_i$ is annotated in the target language context $c_i$ such that $a_i$ is a span within $b_{i}$. 
    }
  \label{annotation_figure}
\end{figure*}

To satisfy these desiderata, we identified the method described below and illustrated in Figure~\ref{annotation_figure}. 
Wikipedia represents a convenient textual domain, as its size and multi-linguality enables collection of data in many diverse languages at scale. It has been used to build many existing QA training resources, allowing us to leverage these to train QA models, without needing to build our own training dataset.
We choose English as our source language as it has the largest Wikipedia, and to easily source crowd workers. We choose six other languages which represent a broad range of linguistic phenomena and have sufficiently large Wikipedia.
Our annotation pipeline consists of three main steps:

Step 1) We automatically extract paragraphs which contain a parallel sentence from articles on the same topic in each language (left of Figure~\ref{annotation_figure}).

Step 2) We employ crowd-workers to annotate questions and answer spans on the English paragraphs (centre of Figure \ref{annotation_figure}). Annotators must choose answer spans within the parallel source sentence. This allows annotation of questions in the source language with high probability of being answerable in the target languages, even if the rest of the context paragraphs are different.

Step 3) We employ professional translators to translate the questions and to annotate answer spans in the target language (right of Figure~\ref{annotation_figure}).

The following sections describe each step in the data collection pipeline in more detail.

\subsection{Parallel Sentence Mining}

Parallel Sentence mining allows us to leverage naturally-written documents and avoid translation, which would be expensive and result in potentially unnatural documents.
In order for questions to be answerable in every target language, we use contexts containing an $N$-way parallel sentence.
Our approach is similar to WikiMatrix~\cite{schwenk:2019_arxiv:wikimatrix} which extracts parallel sentences for many language pairs in Wikipedia, but we limit the search for parallel sentences to documents on the same topic only, and aim for $N$-way parallel sentences.

To detect parallel sentences we use the LASER toolkit,\footnote{\url{https://github.com/facebookresearch/LASER}}
which achieves state-of-the-art performance in mining parallel sentences~\cite{artetxe_margin-based_2019}.
LASER uses multilingual sentence embeddings and a distance or margin criterion in the embeddings space to detect parallel sentences.
The reader is referred to \citet{artetxe_massively_2018} and \citet{artetxe_margin-based_2019} for a detailed description.
See Appendix \ref{appendix_mining} for further details and statistics on the number of parallel sentences mined for all language pairs.

\begin{table}[]
    \centering
    \small
    \begin{tabular}{*{7}{c}}
      \toprule
      \textbf{de}   &  \textbf{es}  &   \textbf{ar}  &   \textbf{zh}  &   \textbf{vi} &  \textbf{hi} \\ 
      \midrule
      5.4M & 1.1M & 83.7k & 24.1K & 9.2k & 1340 \\
      \bottomrule
    \end{tabular}
    \caption{Incremental alignment with English to obtain 7-way aligned sentences.}
    \label{TabIncAlign}
\end{table}

We first independently align all languages with English, then intersect these sets of parallel sentences, forming sets of N-way parallel sentences.
As shown in Table~\ref{TabIncAlign}, starting with 5.4M parallel English/German sentences, the number of N-way parallel sentences quickly decreases as more languages are added. We also found that 7-way parallel sentences lack linguistic diversity, and often appear in the first sentence or paragraph of articles. 

As a compromise between language-parallelism and both the number and diversity of parallel sentences, we use sentences that are 4-way parallel. This yields 385,396 parallel sentences
(see Appendix~\ref{appendix_mining})
which were sub-sampled to ensure parallel sentences were evenly distributed in paragraphs. We ensure that each language combination is equally represented, so that each language has many QA instances in common with every other language. Except for any rejected instances later in the pipeline, each QA instance will be parallel between English and three target languages.

\subsection{English QA Annotation}

We use Amazon Mechanical Turk to annotate English QA instances, broadly following the methodology of \citet{rajpurkar_squad:_2016}. We present workers with an English aligned sentence, $b_{en}$ along with the paragraph that contains it $c_{en}$. Workers formulate a question $q_{en}$ and highlight the shortest answer span $a_{en}$ that answers it.  $a_{en}$ must be be a subspan of $b_{en}$ to ensure $q_{en}$ will be answerable in the target languages. We include a ``No Question Possible" button when no sensible question could be asked. Screenshots of the annotation interface can be found in Appendix \ref{ui_appendix}. The first 15 questions from each worker are manually checked, after which the worker is contacted with feedback, or their work is auto-approved.

Once the questions and answers have been annotated, we run another task to re-annotate English answers. Here, workers are presented with $q_{en}$ and $c_{en}$, and requested to generate an $a'_{en}$ or to indicate that $q_{en}$ is not answerable. Two additional answer span annotations are collected for each question.
The additional answer annotations enable us to calculate an inter-annotator agreement (IAA) score. We calculate the mean token F1 score between the three answer annotations, giving an IAA score of 82\%, comparable to the SQuAD v1.1 development set, where this IAA measure is 84\%.

Rather than provide all three answer annotations as gold answers, we select a single representative reference answer. In 88\% of cases, either two or three of the answers exactly matched, so the majority answer is selected. In the remaining cases, the answer with highest F1 overlap with the other two is chosen.  This results both in an accurate answer span, and ensures the English results are comparable to those in the target languages, where only one answer is annotated per question.

We discard instances where annotators marked the question as unanswerable as well as instances where over 50\% of the question appeared as a sub-sequence of the aligned sentence, as these are too easy or of low quality.
Finally, we reject questions where the IAA score was very low ($<$ 0.3) removing a small number of low quality instances. 
To verify we were not discarding challenging but high quality examples in this step, a manual analysis of discarded questions was performed. Of these discarded questions,  38\% were poorly specified, 24\% did not make sense/had no answer, 30\% had poor answers, and only 8\% were high quality challenging questions.

\subsection{Target Language QA Annotation}

We use the One Hour Translation platform to source professional translators to translate the questions from English to the six target languages, and to find answers in the target contexts. We present each translator with the English question $q_{en}$, English answer $a_{en}$, and the context $c_{x}$ 
(containing aligned sentence $b_x$) in target language $x$. 
The translators are only shown the aligned sentence and the sentence on each side (where these exist). This increases the chance of the question being answerable, as in some cases the aligned sentences are not perfectly parallel, without requiring workers to read the entire context $c_x$. 
By providing the English answer we try to minimize cultural and personal differences in the amount of detail in the answer.

We sample $2\%$ of the translated questions for additional review by language experts. Translators that did not meet the quality standards were removed from the translator pool, and their translations were reallocated. By comparing the distribution of  answer lengths relative to the context to the English distribution, some cases were found where some annotators selected very long answers, especially for Chinese. We clarified the instructions with these specific annotators, and send such cases for re-annotation.
We discard instances in target languages where annotators indicate there is no answer in that language. This means some instances are not 4-way parallel. ``No Answer" annotations occurred for 6.6\%-21.9\% of instances (Vietnamese and German, respectively). 
We release the ``No Answer" data separately as an additional resource, but do not consider it in our experiments or analysis.

\subsection{The Resulting \dataset{} corpus}

\begin{figure*}
\centering

     \begin{subfigure}
         \centering
         \includegraphics[width=0.49\textwidth]{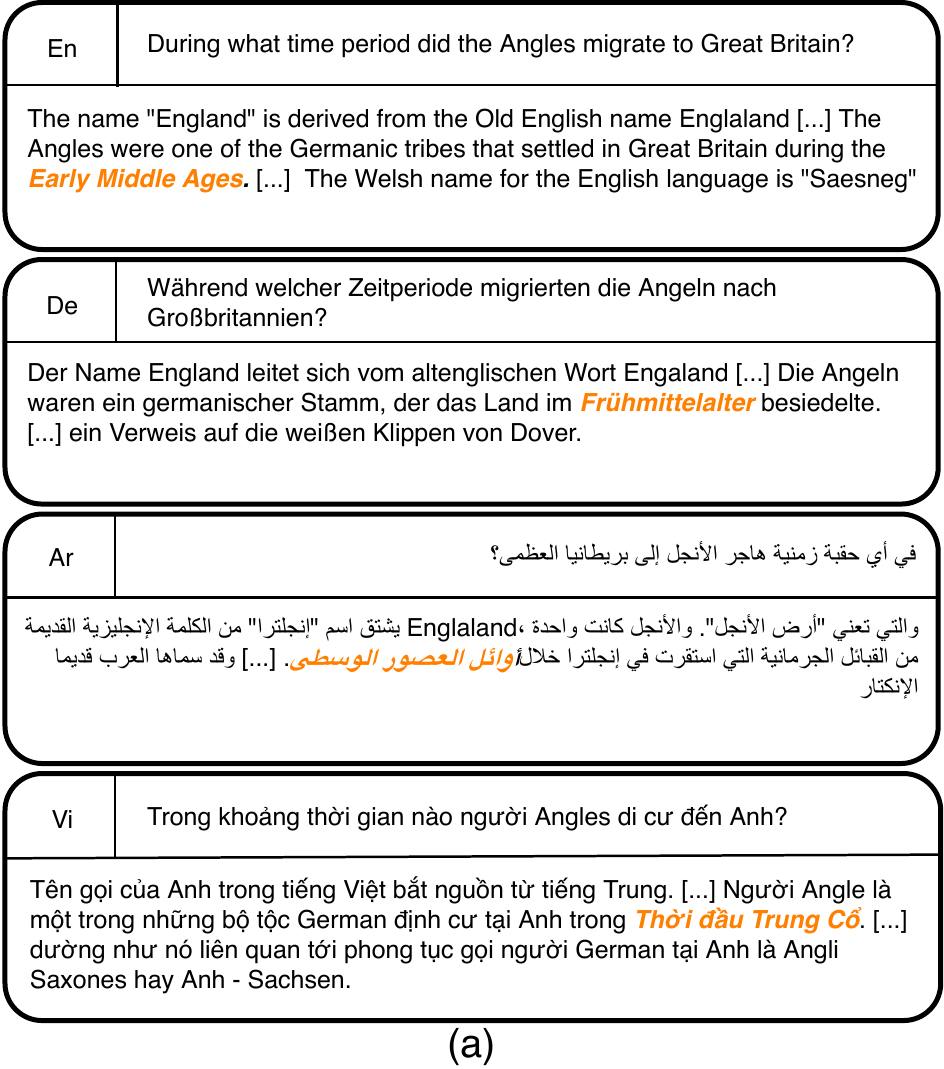}
     \end{subfigure}
     \hfill
     \begin{subfigure}
         \centering
          \includegraphics[width=0.487\textwidth]{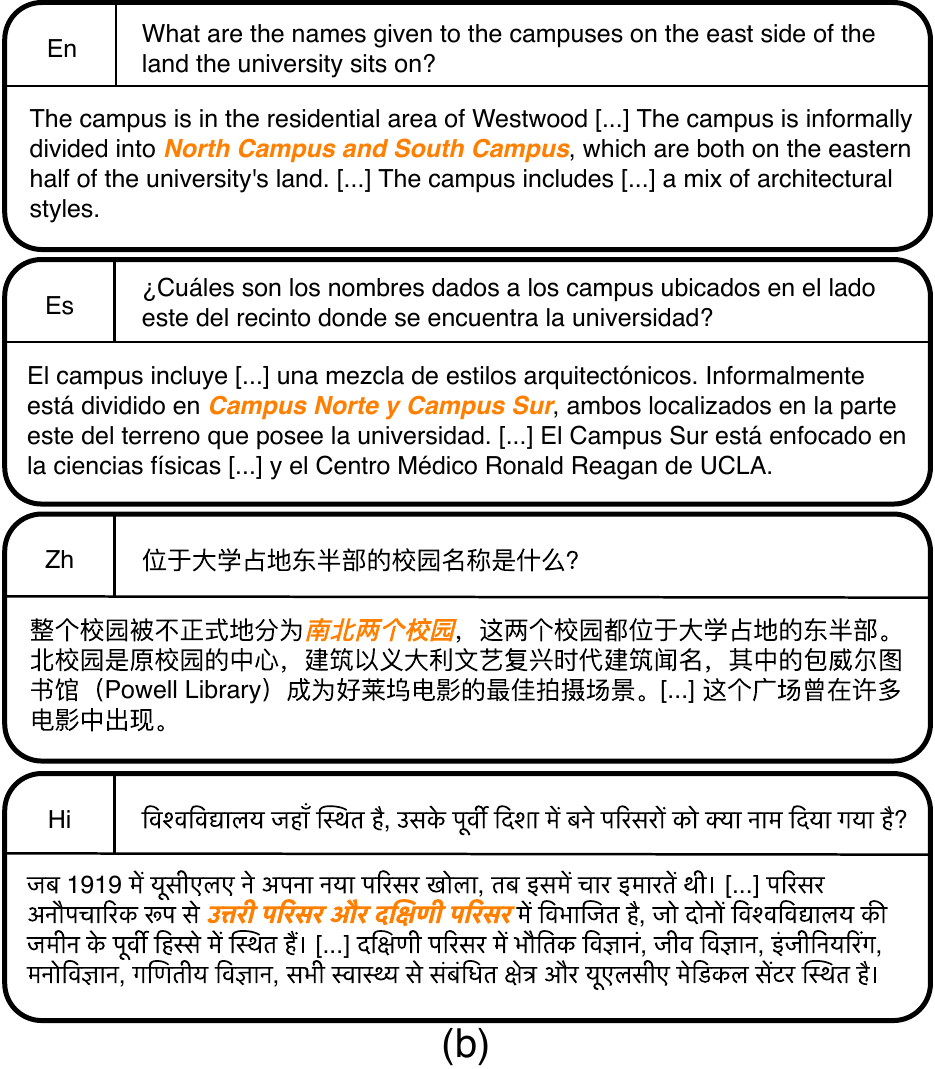}
     \end{subfigure}
\caption{
 (a) \dataset{} example parallel for En-De-Ar-Vi. (b) \dataset{} example parallel for En-Es-Zh-Hi. Answers shown as highlighted spans in contexts. Contexts shortened for clarity with ``[...]". 
}
  \label{dataset_examples}
\end{figure*}

Contexts, questions and answer spans for all the languages are then brought together to create the final corpus. \dataset{} consists of 12,738 extractive QA instances in English and between 5,029 and 6,006 instances in the target languages. 9,019 instances are 4-way parallel, 2,930 are 3-way parallel and 789 2-way parallel. Representative examples are shown in Figure \ref{dataset_examples}.
\dataset{} is split into development and test splits, with statistics in Tables~\ref{stats}, \ref{pairs_fig} and \ref{paragraph_and_article_counts}. 
To investigate the distribution of topics in \dataset{}, a random sample of 500 articles were manually analysed. Articles cover a broad range of topics across different cultures, world regions and disciplines. 23\% are about people, 19\% on physical places, 13\% on cultural topics, 12\% on science/engineering, 9\% on organisations, 6\% on events and 18\% on other topics. Further statistics are given in Appendix \ref{additional_datset_stats}.

\begin{table}[t]
    \centering
    \small
    \begin{tabular}{p{0.5cm}p{0.5cm}p{0.5cm}p{0.5cm}p{0.5cm}p{0.5cm}p{0.5cm}p{0.5cm}p{0.5cm}}
      \toprule
      \textbf{fold} & \textbf{en} & \textbf{de}   &  \textbf{es}  &   \textbf{ar}  &   \textbf{zh}  &   \textbf{vi} &  \textbf{hi} \\ 
      \midrule
      dev & 1148 & 512& 500 & 517 & 504 & 511 & 507 \\
      test & 11590 & 4517 & 5253 & 5335 & 5137 & 5495 & 4918 \\
      \bottomrule
    \end{tabular}
    \caption{Number of instances per language in \dataset{}.}
    \label{stats}
\end{table}

\begin{table}[t]
    \centering
    \small
        \begin{tabular}{ccccccc}
      \toprule
       & \textbf{de}   &  \textbf{es}  &   \textbf{ar}  &   \textbf{zh}  &   \textbf{vi} &  \textbf{hi} \\ 
      \midrule
      \textbf{de}  & 5029 &  &  &  &  &  \\
      \textbf{es}  & 1972 & 5753 &  &  &  &  \\
      \textbf{ar}  & 1856 & 2139 & 5852 &  &  &  \\
      \textbf{zh}  & 1811 & 2108 & 2100 & 5641 &  &  \\
      \textbf{vi}  & 1857 & 2207 & 2210 & 2127 & 6006 &  \\
      \textbf{hi}  & 1593 & 1910 & 2017 & 2124 & 2124 & 5425 \\
      \bottomrule
    \end{tabular}
    \caption{ Number of parallel instances between target language pairs (all instances are parallel with English).}
    \label{pairs_fig}
\end{table}

\begin{table}[h]
    \centering
    \small
    \begin{tabular}{p{1.4cm}p{0.5cm}p{0.4cm}p{0.4cm}p{0.4cm}p{0.4cm}p{0.4cm}p{0.4cm}p{0.4cm}}
      \toprule
          & \hfil\textbf{en} &  \hfil\textbf{de}   &   \hfil\textbf{es}  &   \hfil  \hfil\textbf{ar}  &    \hfil\textbf{zh} &    \hfil\textbf{vi} &   \hfil\textbf{hi} \\ 
      \midrule
      \# Articles  & 5530 & 2806 & 2762 & 2627 & 2673 & 2682& 2255 \\
      \# Contexts & 10894 &4509 & 5215 & 5085 & 4989 & 5246 & 4524 \\
      \# Instances  & 12738 & 5029 & 5753 & 5852 & 5641  & 6006  & 5425  \\
      \bottomrule
    \end{tabular}
    \caption{Number of Wikipedia articles with a context in \dataset{}.}
    \label{paragraph_and_article_counts}
\end{table}

\section{Related Work}
\label{related}

\paragraph{Monolingual QA Data} There is a great variety of English QA data, popularized by MCTest \citep{richardson_mctest:_2013}, CNN/Daily Mail~\citep{hermann_teaching_2015} CBT~\citep{hill_goldilocks_2016}, and WikiQA~\cite{yang_wikiqa:_2015} amongst others. Large span-based datasets such as SQuAD~\citep{rajpurkar_squad:_2016,rajpurkar_know_2018}, TriviaQA~\citep{joshi_triviaqa:_2017}, NewsQA~\citep{trischler_newsqa:_2017}, and Natural Questions~\citep{kwiatkowski_natural_2019} have seen extractive QA become a dominant paradigm. 
However, large, high-quality datasets in other languages are relatively rare. There are several Chinese datasets, such as DUReader~\cite{he_dureader:_2018}, CMRC~\cite{cui_span-extraction_2019} and DRCD~\cite{shao_drcd:_2018}. More recently, there have been efforts to build corpora in a wider array of languages, such as Korean~\cite{korquad} and Arabic~\citep{mozannar_neural_2019}.

\paragraph{Cross-lingual QA Modelling} Cross-lingual QA as a discipline has been explored in QA for RDF data for a number of years, such as the QALD-3 and 5 tracks~\cite{cimiano_multilingual_2013,unger_question_2015}, with more recent work from \citet{zimina_mug-qa:_2018}. \citet{lee_semi-supervised_2018} explore an approach to use English QA data from SQuAD to improve QA performance in Korean  using an in-language seed dataset. \citet{kumar_cross-lingual_2019} study question generation by leveraging English questions to generate better Hindi questions, and \citet{lee_cross-lingual_2019} and \citet{cui_cross-lingual_2019} develop modelling approaches to improve performance on Chinese QA tasks using English resources. \citet{lee_learning_2019} and \citet{hsu_zero-shot_2019} explore modelling approaches for zero-shot transfer and \citet{singh_xlda:_2019} explore how training with cross-lingual data regularizes QA models.

\paragraph{Cross-lingual QA Data} \citet{gupta_mmqa:_2018} release a parallel QA dataset in English and Hindi, \citet{hardalov_beyond_2019} investigate QA transfer from English to Bulgarian, \citet{liu_xcmrc:_2019} release a cloze QA dataset in Chinese and English, and \citet{jing_bipar:_2019} released BiPar, built using parallel paragraphs from novels in English and Chinese. These datasets have a similar spirit to \dataset{}, but are limited to two languages. \citet{asai_multilingual_2018} investigate extractive QA on a manually-translated set of 327 SQuAD instances in Japanese and French, and develop a phrase-alignment modelling technique, showing improvements over back-translation. Like us, they build multi-way parallel extractive QA data, but \dataset{} has many more instances, covers more languages and does not require manual document translation.
\citet{liu_xqa:_2019} explore cross-lingual open-domain QA with a dataset built from Wikipedia ``Did you know?" questions, covering nine languages. Unlike \dataset{}, it is distantly supervised, the dataset size varies by language, instances are not parallel, and answer distributions vary by language, making quantitative comparisons across languages challenging. Finally, in contemporaneous work, \citet{artetxe_cross-lingual_2019} release XQuAD, a dataset of 1190 SQuAD instances from 240 paragraphs manually translated into 10 languages. As shown in Table \ref{paragraph_and_article_counts}, \dataset{} covers 7 languages, but contains more data per language -- over 5k QA pairs from \~5k paragraphs per language. \dataset{} also uses real Wikipedia contexts rather than manual translation. 

\paragraph{Aggregated Cross-lingual Benchmarks} Recently, following the widespread adoption of projects such as GLUE~\citep{wang2018glue}, there have been efforts to compile a suite of high quality multilingual tasks as a unified benchmark system. Two such projects, XGLUE \cite{Liang2020XGLUEAN} and XTREME~\cite{Hu2020XTREMEAM} incorporate \dataset{} as part of their aggregated benchmark.

\section{Cross-lingual QA Experiments}

We introduce two tasks to assess cross-lingual QA performance with \dataset{}.
The first, \emph{cross-lingual transfer}~(XLT), requires training a model with $(c_x, q_x, a_x)$ training data in language $x$, in our case English. Development data in language $x$ is used for tuning. At test time, the model must extract answer $a_y$ in language $y$ given context $c_y$ and question $q_y$.
The second task, \emph{generalized cross-lingual transfer}~(G-XLT), is trained in the same way,  but at test time the model must extract $a_z$ from $c_z$ in language $z$ given $q_y$ in language $y$. This evaluation setup is possible because \dataset{} is highly parallel, allowing us to swap $q_z$ for $q_y$ for parallel instances without changing the question's meaning.

As \dataset{} only has development and test data, we adopt SQuAD v1.1 as training data. We use \dataset{}-en as development data, and focus on zero-shot evaluation, where no training or development data is available in target languages.
Models were trained with the SQuAD-v1 training method from \citet{devlin_bert:_2019} and implemented in Pytext~\cite{aly2018pytext}.
We establish a number of baselines to assess current cross-lingual QA capabilities:

\paragraph{Translate-Train} We translate instances from the SQuAD training set into the target language using machine-translation.\footnote{We use Facebook's production  translation models.}  Before translating, we enclose answers in quotes, as in \citet{lee_semi-supervised_2018}.  This makes it easy to extract answers from translated contexts, and encourages the translation model to map answers into single spans.  We discard instances where this fails (${\sim}5$\%).  This corpus is then used to train a model in the target language.

\paragraph{Translate-Test} The context and question in the target language is translated into English at test time.  We use our best English model to produce an answer span in the translated paragraph.  For all languages other than Hindi,\footnote{Alignments were unavailable for Hindi-English due to production model limitations. Instead we translate English answers using another round of translation. Back-translated answers may not map back to spans in the original context, so this Translate-Test performs poorly.}
we use attention scores, $a_{ij}$, from the translation model to map the answer back to the original language.  Rather than aligning spans by attention argmax, as by \citet{asai_multilingual_2018}, we identify the span in the original context which maximizes F1 score with the English span:
\begin{equation}
\label{f1align}
\begin{split}
\text{RC} &= \textstyle \sum_{i \in S_e, j \in S_o}{a_{ij}} \big/  \sum_{i \in S_e} a_{i*}  \\
\text{PR} &= \textstyle \sum_{i \in S_e, j \in S_o}{a_{ij}} \big/ \sum_{j \in S_o} a_{*j} \\
\text{F1} &= 2 * \text{RC} * \text{PR} \big/  \text{RC} + \text{PR} \\
\text{answer} &= \argmax_{S_o} \text{ F1}(S_o) \\
\end{split}
\end{equation}
where $S_e$ and $S_o$ are the English and original spans respectively, $a_{i*} = \sum_j a_{ij}$ and $a_{*j} = \sum_i a_{*j}$.

\paragraph{Cross-lingual Representation Models} We produce zero-shot transfer results from multilingual BERT (cased, 104 languages) \cite{devlin_bert:_2019} and XLM (MLM + TLM, 15 languages) \cite{lample_cross-lingual_2019}. Models are trained with the SQuAD training set and evaluated directly on the \dataset{} test set in the target language.  Model selection is also constrained to be strictly zero-shot, using only English development data to pick hyper-parameters.  As a result, we end up with a single model that we test for all 7 languages.

\subsection{Evaluation Metrics for Multilingual QA}
\label{eval_metrics}

Most extractive QA tasks use Exact Match (EM) and mean token F1 score as performance metrics.  The widely-used SQuAD evaluation also performs the following answer-preprocessing operations:  i) lowercasing, ii) stripping (ASCII) punctuation iii) stripping (English) articles and iv) whitespace tokenisation. We introduce the following modifications for fairer multilingual evaluation: Instead of stripping ASCII punctuation, we strip all unicode characters with a punctuation \emph{General\_Category}.\footnote{\url{http://www.unicode.org/reports/tr44/tr44-4.html\#General\_Category\_Values}} When a  language has stand-alone articles (English, Spanish, German and Vietnamese) we strip them. We use whitespace tokenization for all \dataset{} languages other than Chinese, where we use the mixed segmentation method from ~\citet{cui_span-extraction_2019}.

\section{Results}

\subsection{XLT Results}
Table \ref{cross_lingual_table} shows the results on the XLT task. XLM performs best overall, transferring best in Spanish, German and Arabic, and competitively with translate-train+M-BERT for Vietnamese and Chinese. XLM is however, weaker in English. Even for XLM, there is a 39.8\% drop in mean EM score (20.9\% F1) over the English BERT-large baseline, showing significant room for improvement. All models generally struggle on Arabic and Hindi.

\begin{table*}[h!]
\small
  \centering
    \begin{tabular}[t]{p{3.13cm}p{1.35cm}p{1.35cm}p{1.35cm}p{1.35cm}p{1.47cm}p{1.35cm}p{1.35cm} }
    \toprule
     F1 / EM  & \multicolumn{1}{c}{\textbf{en}} & \multicolumn{1}{c}{\textbf{es}} & \multicolumn{1}{c}{\textbf{de}} & \multicolumn{1}{c}{\textbf{ar}} & \multicolumn{1}{c}{\textbf{hi}} & \multicolumn{1}{c}{\textbf{vi}} & \multicolumn{1}{c}{\textbf{zh}} \\
     \midrule
     BERT-Large         & \textbf{80.2 / 67.4} & - &- &- &- &- &-\\
     Multilingual-BERT  & 77.7 / 65.2 & 64.3 / 46.6 & 57.9 / 44.3& 45.7 / 29.8 & 43.8 / 29.7 & 57.1 / 38.6 & 57.5 / 37.3\\
     XLM                & 74.9 / 62.4 & \textbf{68.0 / 49.8} & \textbf{62.2 / 47.6} & \textbf{54.8 / 36.3} & 48.8 / 27.3 & 61.4 / 41.8 & 61.1 / \textbf{39.6}\\
     \midrule
     Translate test, BERT-L    & - & 65.4 / 44.0 & 57.9 / 41.8 & 33.6 / 20.4 & 23.8 / 18.9$^{*}$ & 58.2 / 33.2& 44.2 / 20.3\\
     Translate train, M-BERT   & - & 53.9 / 37.4 & 62.0 / 47.5 & 51.8 / 33.2 & \textbf{55.0 / 40.0} & \textbf{62.0 / 43.1} & \textbf{61.4} / 39.5\\
     Translate train, XLM      & -  & 65.2 / 47.8 & 61.4 / 46.7 & 54.0 / 34.4 & 50.7 / 33.4 & 59.3 / 39.4 & 59.8 / 37.9 \\
     \bottomrule
    \end{tabular}
\caption{F1 score and Exact Match on the \dataset{} test set for the cross-lingual transfer task (XLT)}
\label{cross_lingual_table}
\end{table*}

A manual analysis of cases where XLM failed to exactly match the gold answer was carried out for all languages. 39\% of these errors were completely wrong answers, 5\% were annotation errors and 7\% were acceptable answers with no overlap with the gold answer. The remaining 49\% come from answers that partially overlap with the gold span. The variation of errors across languages was small.

\begin{figure}
\centering
  \includegraphics[width=0.48\textwidth]{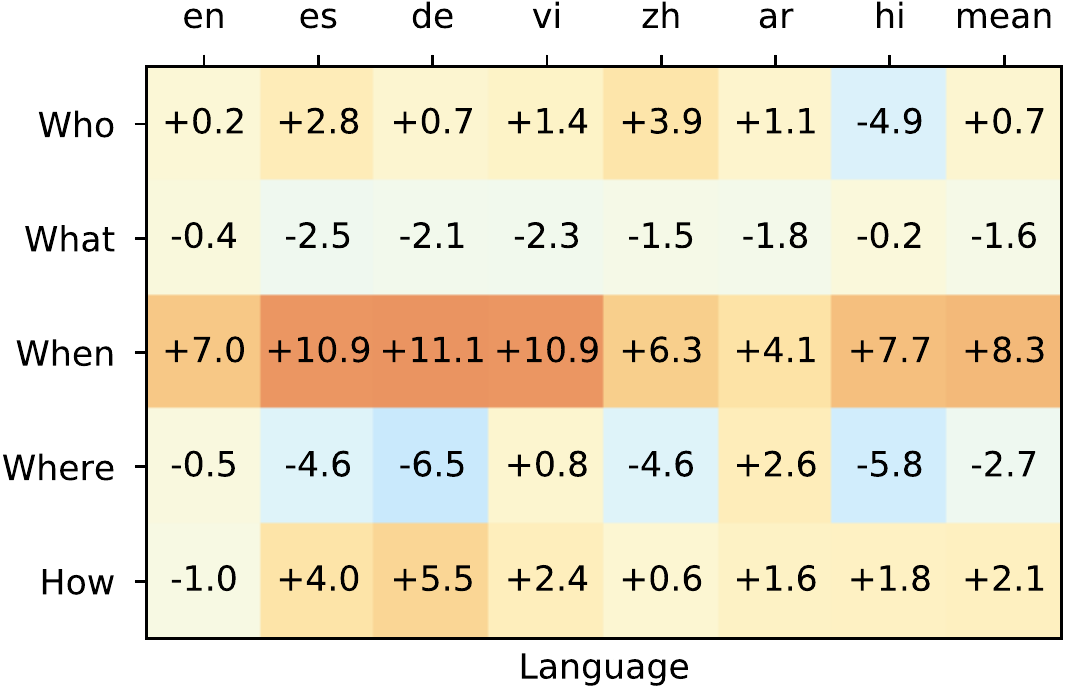}
  \caption{F1 score stratified by English wh* word, relative to overall F1 score for XLM}
  \label{wh_word_performance}
\end{figure}

To see how performance varies by question type, we compute XLM F1 scores stratified by common English wh-words. Figure \ref{wh_word_performance} shows that ``When" questions are the easiest for all languages, and ``Where" questions seem challenging in most target languages. Further details are in Appendix \ref{stratification_appendix}.

\begin{figure}
\centering
  \includegraphics[width=0.48\textwidth]{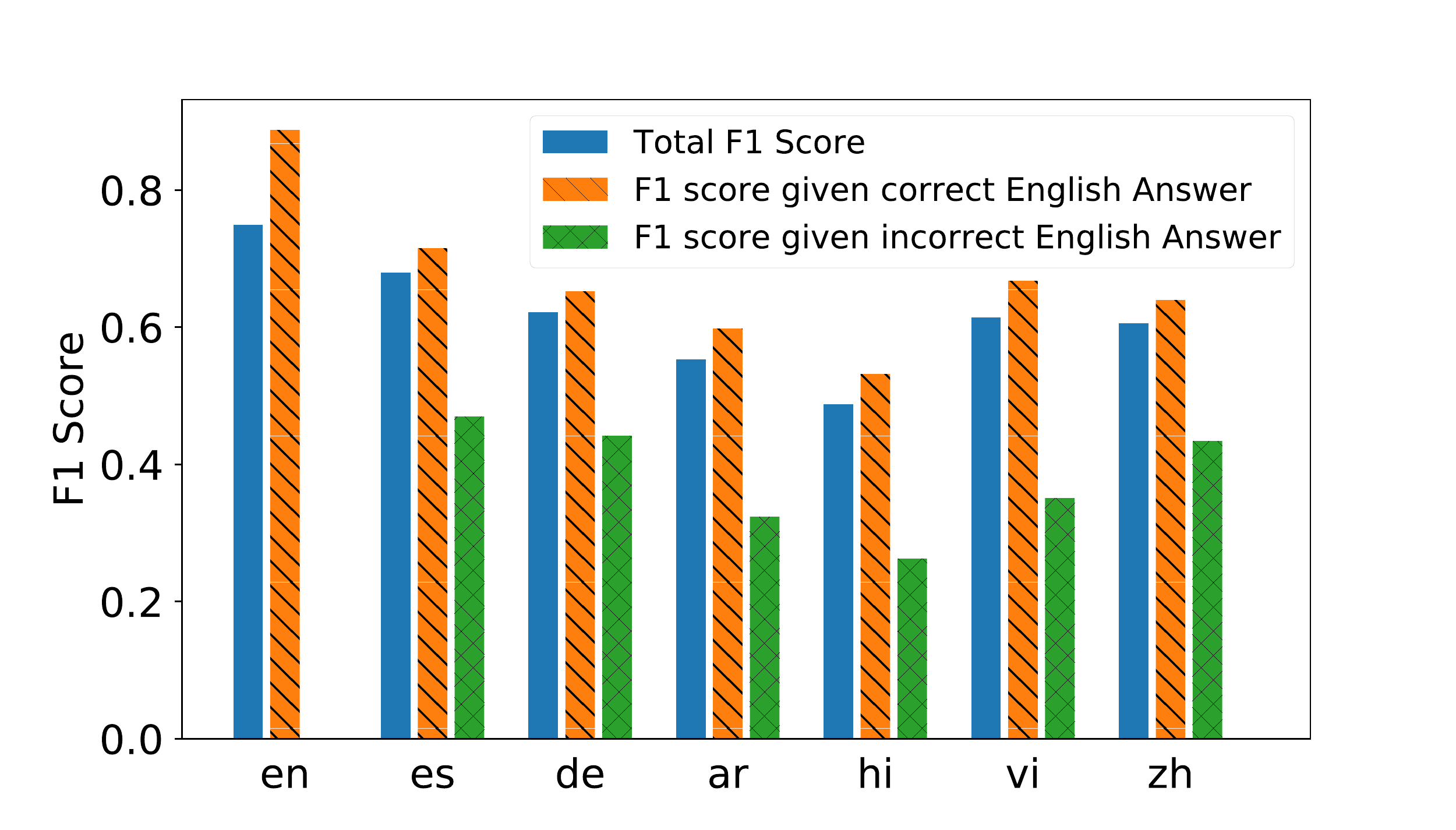}
  \caption{XLM F1 score stratified by English difficulty}
  \label{hard_examples_per_lang}
\end{figure}

To explore whether questions that were difficult for the model in English were also challenging in the target languages, we split \dataset{} into two subsets on whether the XLM model got an English F1 score of zero. 
Figure \ref{hard_examples_per_lang} shows that transfer performance is better when the model answers well in English, but is far from zero when the English answer is wrong, suggesting some questions may be easier to answer in some languages than others.

\subsection{G-XLT Results}

Table~\ref{gxlt_xlm_table} shows results for XLM on the G-XLT task.\footnote{Additional results may be found in Appendix \ref{appendix-gxlt}}
For questions in a given language, the model performs best when the context language matches the question, except for Hindi and Arabic. For contexts in a given language, English questions tend to perform best, apart from Chinese and Vietnamese.

\begin{table}[h]
\small
\centering
%\begin{tabular}[t]{l*{7}{r}r}
\begin{tabular}{p{0.25cm}p{0.55cm}p{0.55cm}p{0.55cm}p{0.55cm}p{0.55cm}p{0.55cm}p{0.55cm}}
\toprule
c/q & \textbf{en} & \textbf{es} & \textbf{de} & \textbf{ar} & \textbf{hi} & \textbf{vi} & \textbf{zh}\\
 \midrule\textbf{en}  & \cellcolor[rgb]{0.463875,0.463875,0.463875}  74.9  & \cellcolor[rgb]{0.588000,0.588000,0.588000}  65.0  & \cellcolor[rgb]{0.668125,0.668125,0.668125}  58.5  & \cellcolor[rgb]{0.764750,0.764750,0.764750}  50.8  & \cellcolor[rgb]{0.854500,0.854500,0.854500}  43.6  & \cellcolor[rgb]{0.703875,0.703875,0.703875}  55.7  & \cellcolor[rgb]{0.726500,0.726500,0.726500}  53.9 \\
\textbf{es}  & \cellcolor[rgb]{0.530625,0.530625,0.530625}  69.5  & \cellcolor[rgb]{0.550375,0.550375,0.550375}  68.0  & \cellcolor[rgb]{0.629000,0.629000,0.629000}  61.7  & \cellcolor[rgb]{0.725000,0.725000,0.725000}  54.0  & \cellcolor[rgb]{0.781625,0.781625,0.781625}  49.5  & \cellcolor[rgb]{0.674125,0.674125,0.674125}  58.1  & \cellcolor[rgb]{0.693125,0.693125,0.693125}  56.5 \\
\textbf{de}  & \cellcolor[rgb]{0.518000,0.518000,0.518000}  70.6  & \cellcolor[rgb]{0.553250,0.553250,0.553250}  67.7  & \cellcolor[rgb]{0.622750,0.622750,0.622750}  62.2  & \cellcolor[rgb]{0.682000,0.682000,0.682000}  57.4  & \cellcolor[rgb]{0.776000,0.776000,0.776000}  49.9  & \cellcolor[rgb]{0.648125,0.648125,0.648125}  60.1  & \cellcolor[rgb]{0.683625,0.683625,0.683625}  57.3 \\
\textbf{ar}  & \cellcolor[rgb]{0.649750,0.649750,0.649750}  60.0  & \cellcolor[rgb]{0.677625,0.677625,0.677625}  57.8  & \cellcolor[rgb]{0.714125,0.714125,0.714125}  54.9  & \cellcolor[rgb]{0.714750,0.714750,0.714750}  54.8  & \cellcolor[rgb]{0.869500,0.869500,0.869500}  42.4  & \cellcolor[rgb]{0.768250,0.768250,0.768250}  50.5  & \cellcolor[rgb]{0.856000,0.856000,0.856000}  43.5 \\
\textbf{hi}  & \cellcolor[rgb]{0.654750,0.654750,0.654750}  59.6  & \cellcolor[rgb]{0.696750,0.696750,0.696750}  56.3  & \cellcolor[rgb]{0.768875,0.768875,0.768875}  50.5  & \cellcolor[rgb]{0.844625,0.844625,0.844625}  44.4  & \cellcolor[rgb]{0.790125,0.790125,0.790125}  48.8  & \cellcolor[rgb]{0.788625,0.788625,0.788625}  48.9  & \cellcolor[rgb]{0.896875,0.896875,0.896875}  40.2 \\
\textbf{vi}  & \cellcolor[rgb]{0.646875,0.646875,0.646875}  60.2  & \cellcolor[rgb]{0.654625,0.654625,0.654625}  59.6  & \cellcolor[rgb]{0.735375,0.735375,0.735375}  53.2  & \cellcolor[rgb]{0.791000,0.791000,0.791000}  48.7  & \cellcolor[rgb]{0.893625,0.893625,0.893625}  40.5  & \cellcolor[rgb]{0.632625,0.632625,0.632625}  61.4  & \cellcolor[rgb]{0.793250,0.793250,0.793250}  48.5 \\
\textbf{zh}  & \cellcolor[rgb]{0.738375,0.738375,0.738375}  52.9  & \cellcolor[rgb]{0.702500,0.702500,0.702500}  55.8  & \cellcolor[rgb]{0.775625,0.775625,0.775625}  50.0  & \cellcolor[rgb]{0.888875,0.888875,0.888875}  40.9  & \cellcolor[rgb]{0.957250,0.957250,0.957250}  35.4  & \cellcolor[rgb]{0.818875,0.818875,0.818875}  46.5  & \cellcolor[rgb]{0.636750,0.636750,0.636750}  61.1 \\
\bottomrule
\end{tabular}

\caption{F1 Score for XLM for G-XLT. Columns show question language, rows show context language. }
\label{gxlt_xlm_table}
\end{table}

\subsection{English Results on SQuAD 1 and \dataset{}}
\label{english_comparison_section}
The \dataset{}-en results in Table \ref{cross_lingual_table} are lower than reported results on SQuAD v1.1 in the literature for equivalent models. However, once SQuAD scores are adjusted to reflect only having one answer annotation (picked using the same method used to pick \dataset{} answers), the discrepancy drops to 5.8\% on average (see Table \ref{english_comparisons}). \dataset{}-en contexts are on average 28\% longer than SQuAD's, and \dataset{} covers a much wider set of articles than SQuAD.
Minor differences in preprocessing and answer lengths may also contribute (\dataset{}-en answers are slightly longer, 3.1 tokens vs 2.9 on average).  Question type distributions are very similar in both datasets (Figure \ref{wh_distro} in Appendix \ref{appendix_a})

\begin{table}
\small
    \centering
    \begin{tabular}{llcc}
    \toprule
        \textbf{Model} & \textbf{SQuAD} & \textbf{SQuAD*} & \textbf{\dataset{}}-en \\
        \midrule
        BERT-Large & 91.0 / 80.8 & 84.8 / 72.9 & 80.2 / 67.4 \\
        M-BERT & 88.5 / 81.2 & 83.0 / 71.1  & 77.7	/ 65.1\\
        XLM & 87.6 / 80.5 & 82.1 / 69.7 & 74.9 / 62.4 \\
         \bottomrule
    \end{tabular}
    \caption{English performance comparisons to SQuAD using our models. * uses a single answer annotation.}
    \label{english_comparisons}
\end{table}

\section{Discussion}

It is worth discussing the quality of context paragraphs in \dataset{}. Our parallel sentence mining approach can source independently-written documents in different languages, but, in practice, articles are often translated from English to the target languages by volunteers. Thus our method sometimes acts as an efficient mechanism of sourcing existing human translations, rather than  sourcing independently-written content on the same topic. The use of machine translation is strongly discouraged by the Wikipedia community,\footnote{\url{https://en.wikipedia.org/wiki/Wikipedia:Translation\#Avoid_machine_translations}} but from examining edit histories of articles in \dataset{}, machine translation is occasionally used as an article seed, before being edited and added to by human authors. 

Our annotation method restricts answers to come from specified sentences. Despite being provided several sentences of context, some annotators may be tempted to only read the parallel sentence and write questions which only require a single sentence of context to answer. However, single sentence context questions are a known issue in SQuAD annotation in general~\cite{sugawara_what_2018} suggesting our method would not result in less challenging questions, supported by scores on \dataset{}-en being similar to SQuAD (section \ref{english_comparison_section}).

\dataset{} is partitioned into development and test splits. As \dataset{} is parallel, this means there is development data for every language. Since \dataset{} will be freely available, this was done to reduce the risk of test data over-fitting in future, and to establish standard splits. However, in our experiments, we only make use of the English development data and study strict zero-shot settings. Other evaluation setups could be envisioned, e.g. by exploiting the target language development sets for hyper-parameter optimisation or fine-tuning, which could be fruitful for higher transfer performance, but we leave such ``few-shot" experiments as future work. Other potential areas to explore involve training datasets other than English, such as CMRC~\cite{cui_span-extraction_2018}, or using unsupervised QA techniques to assist transfer~\cite{lewis_unsupervised_2019}.

Finally, a large body of work suggests QA models are over-reliant on word-matching between question and context \cite{jia_adversarial_2017,gan_improving_2019}. G-XLT represents an interesting test-bed, as simple symbolic matching is less straightforward when questions and contexts use different languages. However, the performance drop from XLT is relatively small (8.2 mean F1), suggesting word-matching in cross-lingual models is more nuanced and robust than it may initially appear.

\section{Conclusion}

We have introduced \dataset{}, a highly-parallel multilingual QA benchmark in seven languages. We developed several baselines on two cross-lingual understanding tasks on \dataset{} with state-of-the-art methods, and demonstrate significant room for improvement. We hope that \dataset{} will help to catalyse work in cross-lingual QA to close the gap between training and testing language performance.

\section{Acknowledgements}

The authors would like to acknowledge their crowd-working and translation colleagues for their work on \dataset{}. The authors would also like to thank Yuxiang Wu, Andres Compara Nu\~nez, Kartikay Khandelwal, Nikhil Gupta, Chau Tran, Ahmed Kishky, Haoran Li, Tamar Lavee, Ves Stoyanov and the anonymous reviewers for their feedback and comments.

\bibliography{patrick,addtl}
\bibliographystyle{acl_natbib}

\clearpage

\twocolumn[
\centering
{\Large \emph{Supplementary Materials for}}\\
\smallskip
{\Large \papername{}}\\
\bigskip
]
\appendix

\section{Appendices}
\label{appendix_a}

\subsection{Annotation Interface}
\label{ui_appendix}

\begin{figure}[t!]
\centering
  \includegraphics[width=0.5\textwidth]{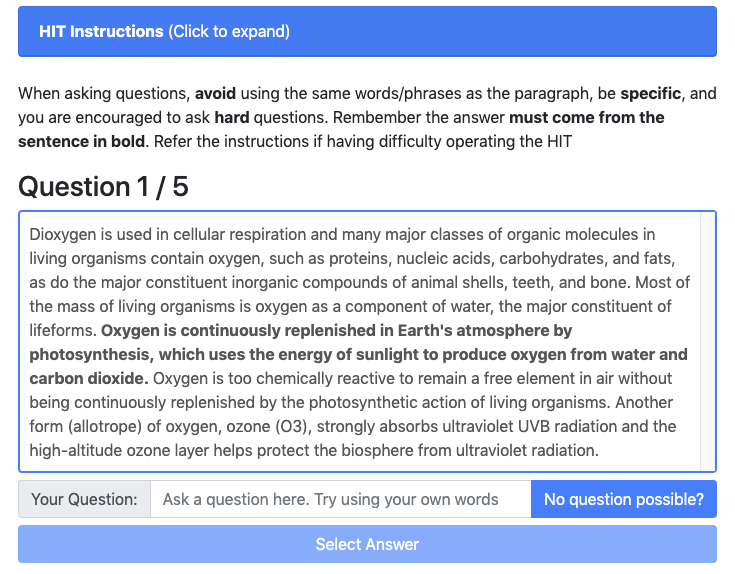}
  \caption{ English QA annotation interface screenshot}
  \label{squad_annotation_figure}
\end{figure}

\begin{figure}[t!]
\centering
  \includegraphics[width=0.5\textwidth]{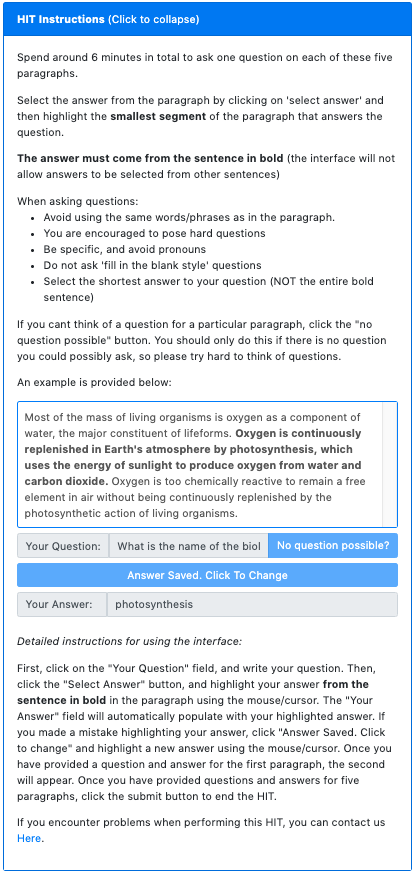}
  \caption{English annotation instructions screenshot}
  \label{annotation_instructions}
\end{figure}

Figure \ref{squad_annotation_figure} shows a screenshot of the annotation interface. Workers are asked to write a question in the box, and highlight an answer using the mouse in the sentence that is in bold. There are a number of data input validation features to assist workers, as well as detailed instructions in a drop-down window, which are shown in Figure \ref{annotation_instructions}

\subsection{Additional \dataset{} Statistics}
\label{additional_datset_stats}
\begin{figure}[h]
\centering
  \includegraphics[width=0.5\textwidth]{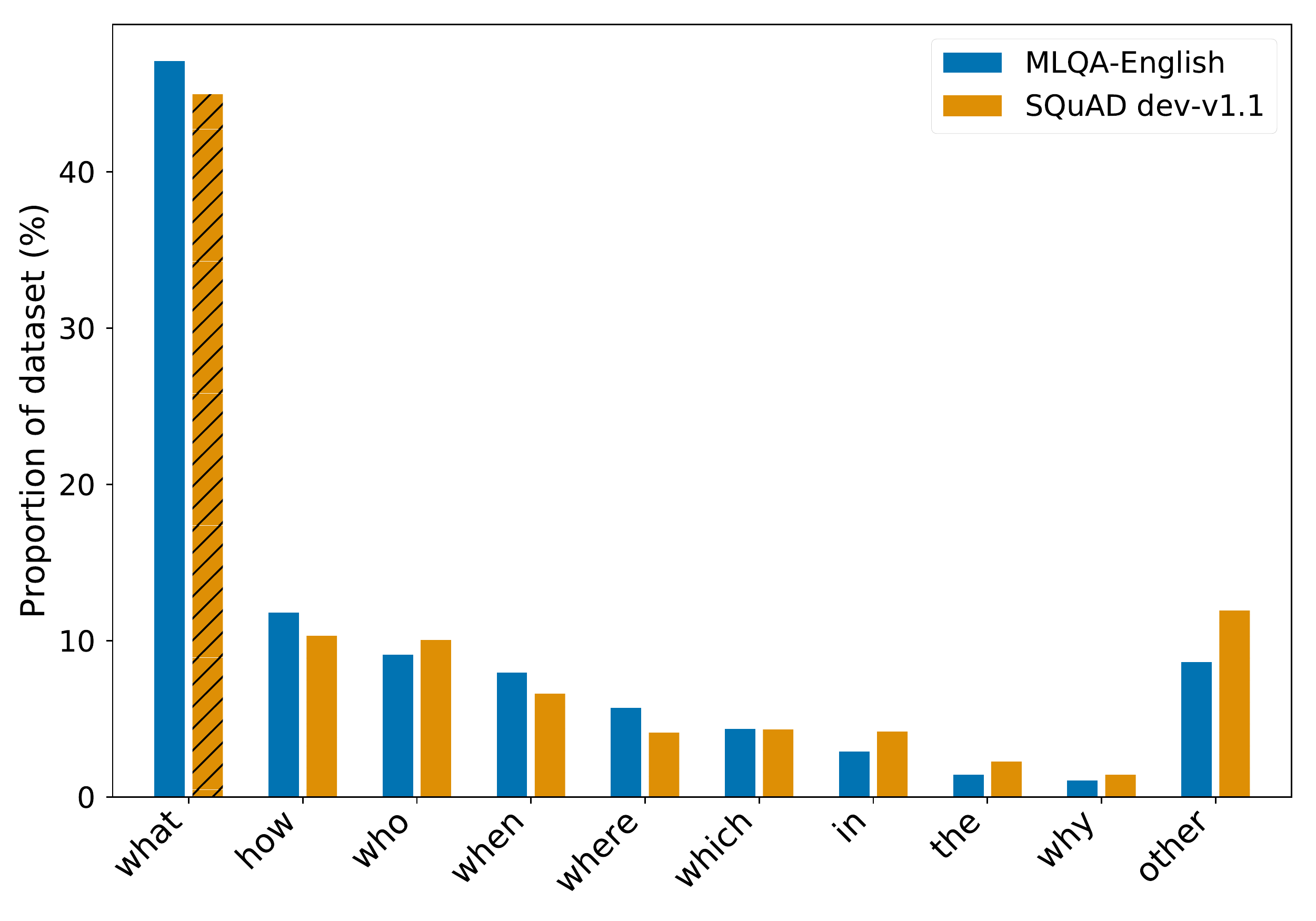}
  \caption{Question type distribution (by ``wh" word) in \dataset{}-en and SQuAD V1.1. The distributions are strikingly similar}
  \label{wh_distro}
\end{figure}

Figure \ref{wh_distro} shows the distribution of wh words in questions in both \dataset{}-en and SQuAD v.1.1. The distributions are very similar, suggesting training on SQuAD data is an appropriate training dataset choice.

Table \ref{paragraph_and_article_counts} shows the number of Wikipedia articles that feature at least one of their paragraphs as a context paragraph in \dataset{}, along with the number of unique context paragraphs in \dataset{}. There are 1.9 context paragraphs from each article on average. This is in contrast to SQuAD, which instead features a small number of curated articles, but more densely annotated, with 43 context paragraphs per article on average. Thus, \dataset{} covers a much broader range of topics than SQuAD.

% \begin{table}[h]
%     \centering
%     \small
%     \begin{tabular}{p{1.4cm}p{0.5cm}p{0.4cm}p{0.4cm}p{0.4cm}p{0.4cm}p{0.4cm}p{0.4cm}p{0.4cm}}
%       \toprule
%       & \hfil\textbf{en} &  \hfil\textbf{ar}   &   \hfil\textbf{de}  &   \hfil  \hfil\textbf{es}  &    \hfil\textbf{hi} &    \hfil\textbf{zh} &   \hfil\textbf{vi} \\ 
%       \midrule
%       \# Articles  &5530 & 2627 & 2806 & 2762 & 2255 & 2673& 2682 \\
%       \# Contexts & 10894 &5085 & 4509 & 5215 & 4524 & 4989 & 5246 \\
%       \# Instances  & 12738 & 5852 & 5029 & 5753 & 5425  & 4989  & 6006  \\
%       \bottomrule
%     \end{tabular}
%     \caption{Number of Wikipedia articles with a context in \dataset{}.}
%     \label{paragraph_and_article_counts}
% \end{table}

\begin{table}
    \centering
    \small
    \begin{tabular}{p{1cm}p{0.45cm}p{0.45cm}p{0.45cm}p{0.45cm}p{0.45cm}p{0.45cm}p{0.45cm}p{0.45cm}}
      \toprule
       & \textbf{en} & \textbf{de}   &  \textbf{es}  &   \textbf{ar}  &   \textbf{zh*}  &   \textbf{vi} &  \textbf{hi} \\ 
      \midrule
      Context  & 157.5 & 102.2 & 103.4 & 116.8 & 222.9 & 195.1 & 141.5 \\
      Question & 8.4 & 7.7 & 8.6 & 7.6 & 14.3 & 10.6 & 9.3 \\
      Answer   & 3.1 & 3.2 & 4.1 & 3.4 & 8.2  & 4.5  & 3.6  \\
      \bottomrule
    \end{tabular}
    \caption{Mean Sequence lengths (tokens) in \dataset{}. *calculated with mixed segmentation (section \ref{eval_metrics})}
    \label{lengths}
\end{table}

Table \ref{lengths} shows statistics about the lengths of contexts, questions and answers in \dataset{}. Vietnamese has the longest contexts on average and German are shortest, but all languages have a substantial tail of long contexts. Other than Chinese, answers are on average 3 to 4 tokens.

\subsection{QA Performance stratified by question and answer types}
\label{stratification_appendix}

\begin{figure}
\centering
  \includegraphics[width=0.5\textwidth]{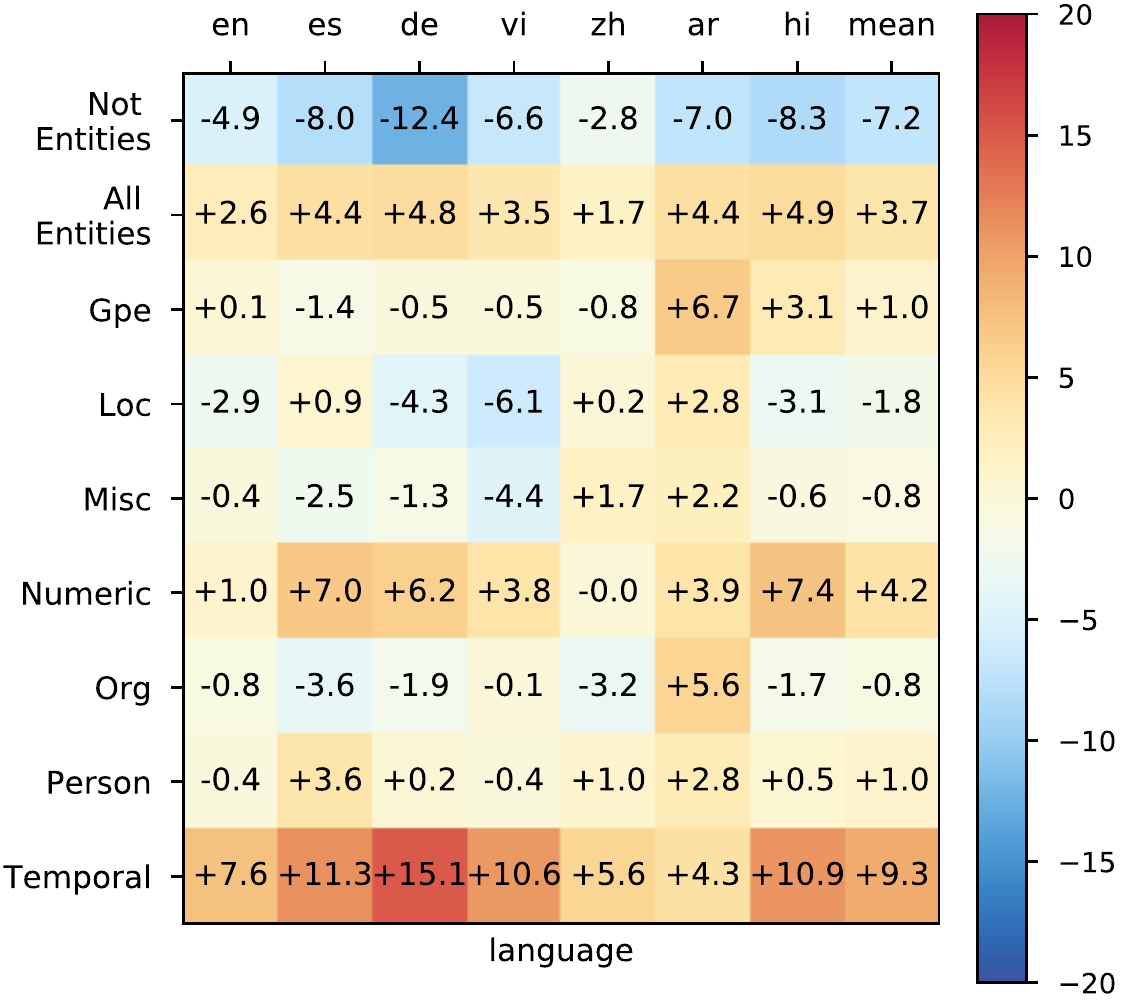}
  \caption{F1 score stratified by named entity types in answer spans, relative to overall F1 score for XLM}
  \label{entity_type_performance}
\end{figure}
To examine how performance varies across languages for different types of questions, we stratify \dataset{} with three criteria --- By English Wh-word, by answer Named-Entity type and by English Question Difficulty

\paragraph{By wh-word:} First, we split by the English Wh* word in the question. This resulting change in F1 score compared to the overall F1 score is shown in Figure \ref{wh_word_performance}, and discussed briefly in the main text. The English wh* word provides a clue as to the type of answer the questioner is expecting, and thus acts as a way of classifying QA instances into types. We chose the 5 most common wh* words in the dataset for this analysis. We see that ``when" questions are consistently easier than average across the languages, but the pattern is less clear for other question types. "Who" questions also seem  easier than average, except for Hindi, where the performance is quite low for these questions. ``How"-type questions (such as ``how much", ``how many" or ``how long" ) are also more challenging to answer than average in English compared to the other languages. ``Where" questions also seem challenging for Spanish, German, Chinese and Hindi, but this is not true for Arabic or Vietnamese.

\paragraph{By Named-Entity type} We create subsets of \dataset{} by detecting which English named entities are contained in the answer span. To achieve this, we run Named Entity Recognition using SPaCy~\cite{spacy2}, and detect where named entity spans overlap with answer spans. The F1 scores for different answer types relative to overall F1 score are shown for various Named Entity types in Figure \ref{entity_type_performance}. There are some clear trends: Answer spans that contain named entities are easier to answer than those that do not (the first two rows) for all the languages, but the difference is most pronounced for German. Secondly,``Temporal" answer types (\texttt{DATE} and \texttt{TIME} entity labels) are consistently easier than average for all languages, consistent with the high scores for ``when" questions in the previous section. Again, this result is most pronounced in German, but is also very strong for Spanish, Hindi, and Vietnamese. Arabic also performs well for \texttt{ORG}, \texttt{GPE} and \texttt{LOC} answer types, unlike most of the other languages. Numeric questions (\texttt{CARDINAL}, \texttt{ORDINAL}, \texttt{PERCENT}, \texttt{QUANTITY} and \texttt{MONEY} entity labels) also seem relatively easy for the model in most languages.

\paragraph{By English Question Difficulty} Here, we split \dataset{} into two subsets, according to whether the XLM model got the question completely wrong (no word overlap with the correct answer). We then evaluated the mean F1 score for each language on the two subsets, with the results shown in Figure \ref{hard_examples_per_lang}. We see that questions that are ``easy" in English also seem to be easier in the target languages, but the drop in performance for the      ``hard" subset is not as dramatic as one might expect. This suggests that not all questions that are hard in English in \dataset{} are hard in the target languages. This could be due to the grammar and morphology of different languages leading to questions being easier or more difficult to answer, but an another factor is that context documents can be shorter in target languages for questions the model struggled to answer correctly in English, effectively making them easier. Manual inspection suggests that whilst context documents are often shorter for when the model is correct in the target language, this effect is not sufficient to explain the difference in performance. 

\subsection{Additional G-XLT results}
\label{appendix-gxlt}

Table~\ref{gxlt_xlm_table} in the main text shows for XLM on the G-XLT task, and Table \ref{gxlt_mbert_table} for Multilingual-BERT respectively. XLM outperforms M-BERT for most language pairs, with a mean G-XLT performance of 53.4 F1 compared to 47.2 F1 (mean of off-diagonal elements of Tables \ref{gxlt_xlm_table} and \ref{gxlt_mbert_table}). Multilingual BERT exhibits more of a preference for English than XLM for G-XLT, and exhibits a bigger performance drop going from XLT to G-XLT (10.5 mean drop in F1 compared to 8.2).

\begin{table}[h]
\small
\centering
%\begin{tabular}[t]{l*{7}{r}r}
\begin{tabular}{p{0.25cm}p{0.55cm}p{0.55cm}p{0.55cm}p{0.55cm}p{0.55cm}p{0.55cm}p{0.55cm}}
\toprule
c/q & \textbf{en} & \textbf{es} & \textbf{de} & \textbf{ar} & \textbf{hi} & \textbf{vi} & \textbf{zh}\\
 \midrule\textbf{en}  & \cellcolor[rgb]{0.428250,0.428250,0.428250}  77.7  & \cellcolor[rgb]{0.595375,0.595375,0.595375}  64.4  & \cellcolor[rgb]{0.616250,0.616250,0.616250}  62.7  & \cellcolor[rgb]{0.828625,0.828625,0.828625}  45.7  & \cellcolor[rgb]{0.898250,0.898250,0.898250}  40.1  & \cellcolor[rgb]{0.747000,0.747000,0.747000}  52.2  & \cellcolor[rgb]{0.722875,0.722875,0.722875}  54.2 \\
\textbf{es}  & \cellcolor[rgb]{0.557000,0.557000,0.557000}  67.4  & \cellcolor[rgb]{0.596625,0.596625,0.596625}  64.3  & \cellcolor[rgb]{0.669250,0.669250,0.669250}  58.5  & \cellcolor[rgb]{0.848750,0.848750,0.848750}  44.1  & \cellcolor[rgb]{0.924125,0.924125,0.924125}  38.1  & \cellcolor[rgb]{0.797875,0.797875,0.797875}  48.2  & \cellcolor[rgb]{0.760625,0.760625,0.760625}  51.1 \\
\textbf{de}  & \cellcolor[rgb]{0.614750,0.614750,0.614750}  62.8  & \cellcolor[rgb]{0.682375,0.682375,0.682375}  57.4  & \cellcolor[rgb]{0.676750,0.676750,0.676750}  57.9  & \cellcolor[rgb]{0.915125,0.915125,0.915125}  38.8  & \cellcolor[rgb]{0.955750,0.955750,0.955750}  35.5  & \cellcolor[rgb]{0.840750,0.840750,0.840750}  44.7  & \cellcolor[rgb]{0.821125,0.821125,0.821125}  46.3 \\
\textbf{ar}  & \cellcolor[rgb]{0.759625,0.759625,0.759625}  51.2  & \cellcolor[rgb]{0.834125,0.834125,0.834125}  45.3  & \cellcolor[rgb]{0.819500,0.819500,0.819500}  46.4  & \cellcolor[rgb]{0.829375,0.829375,0.829375}  45.6  & \cellcolor[rgb]{0.998125,0.998125,0.998125}  32.1  & \cellcolor[rgb]{0.934250,0.934250,0.934250}  37.3  & \cellcolor[rgb]{0.899750,0.899750,0.899750}  40.0 \\
\textbf{hi}  & \cellcolor[rgb]{0.752625,0.752625,0.752625}  51.8  & \cellcolor[rgb]{0.859625,0.859625,0.859625}  43.2  & \cellcolor[rgb]{0.822625,0.822625,0.822625}  46.2  & \cellcolor[rgb]{0.938375,0.938375,0.938375}  36.9  & \cellcolor[rgb]{0.852125,0.852125,0.852125}  43.8  & \cellcolor[rgb]{0.919750,0.919750,0.919750}  38.4  & \cellcolor[rgb]{0.893875,0.893875,0.893875}  40.5 \\
\textbf{vi}  & \cellcolor[rgb]{0.633000,0.633000,0.633000}  61.4  & \cellcolor[rgb]{0.748875,0.748875,0.748875}  52.1  & \cellcolor[rgb]{0.757000,0.757000,0.757000}  51.4  & \cellcolor[rgb]{0.969625,0.969625,0.969625}  34.4  & \cellcolor[rgb]{0.961125,0.961125,0.961125}  35.1  & \cellcolor[rgb]{0.686125,0.686125,0.686125}  57.1  & \cellcolor[rgb]{0.810750,0.810750,0.810750}  47.1 \\
\textbf{zh}  & \cellcolor[rgb]{0.675375,0.675375,0.675375}  58.0  & \cellcolor[rgb]{0.786000,0.786000,0.786000}  49.1  & \cellcolor[rgb]{0.779375,0.779375,0.779375}  49.6  & \cellcolor[rgb]{0.893875,0.893875,0.893875}  40.5  & \cellcolor[rgb]{0.950250,0.950250,0.950250}  36.0  & \cellcolor[rgb]{0.841875,0.841875,0.841875}  44.6  & \cellcolor[rgb]{0.681750,0.681750,0.681750}  57.5 \\
\bottomrule
\end{tabular}
\caption{F1 Score for M-BERT for G-XLT. Columns show question language, rows show context language.
}
\label{gxlt_mbert_table}
\end{table}

\subsection{Additional preprocessing Details}

OpenCC (\url{https://github.com/BYVoid/OpenCC}) is used to convert all Chinese contexts to Simplified Chinese, as wikipedia dumps generally consist of a mixture of simplified and traditional Chinese text.
 
\subsection{Further details on Parallel Sentence mining}
\label{appendix_mining}

\begin{table*}[t]
    \centering
    \begin{tabular}{r*{7}{r}}
    \toprule
     \textbf{N-way} &   \textbf{en} &       \textbf{de} &       \textbf{es} &       \textbf{ar} &       \textbf{zh} &       \textbf{vi} &       \textbf{hi} \\
     \midrule
 2 & 12219436 &  3925542 &  4957438 &  1047977 &  1174359 &   904037 &   210083 \\
 3 &  2143675 &  1157009 &  1532811 &   427609 &   603938 &   482488 &    83495 \\   
 4 &   385396 &   249022 &   319902 &   148348 &   223513 &   181353 &    34050 \\   
 5 &    73918 &    56756 &    67383 &    44684 &    58814 &    54884 &    13151 \\   
 6 &    12333 &    11171 &    11935 &    11081 &    11485 &    11507 &     4486 \\    
 7 &     1340 &     1340 &     1340 &     1340 &     1340 &     1340 &     1340 \\    
    \bottomrule
    \end{tabular}
    \caption{Number of mined parallel sentences as a function of how many languages the sentences are parallel between}
    \label{TabMine}
\end{table*}
Table \ref{TabMine} shows the number of mined parallel sentences found in each language, as function of how many languages the sentences are parallel between. As the number of languages that a parallel sentence is shared between increases, the number of such sentences decreases. When we look for 7-way aligned examples, we only find 1340 sentences from the entirety of the 7 Wikipedia. Additionally, most of these sentences are the first sentence of the article, or are uninteresting. However, if we choose 4-way parallel sentences, there are plenty of sentences to choose from. We sample evenly from each combination of English and 3 of the 6 target languages. This ensures that we have an even distribution over all the target languages, as well as ensuring we have even numbers of instances that will be parallel between target language combinations.

\end{document}